# Physics-Data Driven Machine Learning Based Model : A Hybrid Way for Nonlinear, Dynamic, and Open-loop Identification of IPMC Soft Artificial Muscles

Mohsen Annabestani, *Member, IEEE*, Mohammad Hossein Sayyad, Zahra Meskar, Mehdi Fardmanesh, Senior *Member, IEEE,* and Barbara Mazzolai

*Abstract*— Ionic Polymer Metal Composites (IPMCs) are one of the most preferred choices among biocompatible materials for industrial and biomedical applications. Despite their advantages, some of their drawbacks include non-linear and hysteretic behavior, which complicates the modeling process. In previous works, usually autoregressive models were used to predict the behavior of an IPMC actuator. The main drawback of using an autoregressive model is that it cannot be used in mobile and real-time applications. In this study, we proposed a hybrid analytical intelligent model for an IPMC actuator. The most outstanding feature of this model is its non-autoregressive structure. The hybrid concept proposed in this study can be generalized to various problems other than IPMCs. The structure used in this work comprises an analytical model and a deep neural network, providing a non-linear, dynamic, and non-autoregressive model for the IPMC actuator. Lastly, the average NMSE achieved using the proposed hybrid model is 9.5781e-04 showing a significant drop in the error rate compared to other non-autoregressive structures.

## I. INTRODUCTION

IPMC or ionic polymer-metal composite is a soft smart material that as an imitator of biological muscle has variety of potential applications especially in biomedical systems [1-8]. Among the most important distinguishing features of an IPMC actuator are its low density and also considerable stimulus strain. It is lighter than other similar smart materials, and also it can be easily stimulated in response to low input voltage [9, 10]. It comprises a thin layer of an ion exchanging membrane such as Nafion, Flemion, etc. This membrane is surrounded by two metallic plates like Pt, Pd, etc. [11, 12] (Fig.1). The IPMC membrane is full of anions and cations. Anions are fixed, whereas cations can move freely inside the membrane. When an external voltage is applied to IPMC, the hydrated cations move toward the cathode side, and IPMC bends toward the anode [10]. Furthermore, if the IPMC strip is bent, a low voltage is produced between two electrodes. The produced voltage is highly correlated with the bending characteristics of an IPMC actuator. Hence, IPMC can be considered both a sensor and an actuator [11, 13].

There are two main approaches for modeling the IPMC's behavior. The first approach is analytical modeling, and the second one is predictive identification. In the first category, some works use an equivalent distributed RC circuit for modeling the IPMC [14, 15]. Some other works model the behavior of IPMCs based on multiphysics approaches [16-19]. Also, there are some other analytical methods that are not directly related to the first two groups [20-22]. In the second category, predictive identification, there are several works worth mentioning. This category can be split into classical methods and intelligent methods. For the classical methods, [23] used Box-Jenkins and ARMAX method to identify the dynamic behavior of the IPMC actuator. [24] incorporated linear and non-linear least-squares method by an experimental approach to model the IPMC actuator. In [25], A non-linear autoregressive model was used alongside a moving average model with an external input to represent the dynamic and hysteretic behavior of IPMC. In the category of intelligent methods, [26] used a neural network to obtain the parameters of two Mamdani fuzzy systems, describing a non-linear model for IPMC. In [27], the NARX structure was used alongside fuzzy logic and particle swarm optimization to model IPMC's behavior. Annabestani et al. also presented two other autoregressive methods for intelligent identification of the large deformation behavior of IPMC, first was an ANFIS-NARX paradigm [28] which used the potential of ANFIS neuro-fuzzy inference system and the NARX dynamic structure was able to accurately predict the tip displacement of IPMC. In the second one, a new method for non-uniform deformation and curvature identification of IPMC actuators was presented [29].

Mohsen Annabestani and Barbara Mazzolai are with the Istituto Italiano di Tecnologia, Bioinspired Soft Robotics Lab, 16163, Genova GE, Italy (e-mail: mohsen.annabestani@iit.it ; barbara.mazzolai@iit.it ).

Mohammad Hossein Sayyad, Zahra Meskar, and Mehdi Fardmanesh are with the faculty of electrical engineering, Sharif University of Technology, Tehran, Iran.

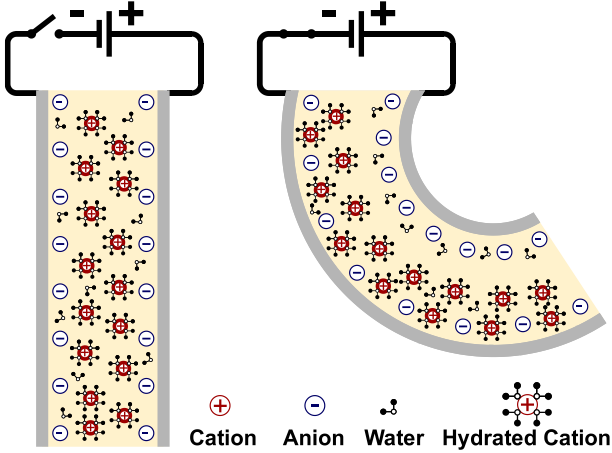

Fig.1. Working principle of an IPMC actuator: (Left) IPMC before applying voltage ; (Right) after applying voltage.

All of the mentioned models are autoregressive, which means they all use the system's output (IPMC displacement or deformation) in their identification process, which makes fundamental constraints for practical applications of IPMCs. For example, in mobile usage of IPMC or in some in vitro applications, we are not able to measure the IPMC tip displacement or output signal online. The problem with online measurements is that they usually cause depreciation of the large bendability of the IPMC actuator. This problem can be solved by using non-autoregressive models. Nonetheless, few works have used non-autoregressive identification. In [30], a network of neural oscillators has been used to model the IPMC in an open-loop structure. In [31], a linear, open-loop model is presented to model the actuation behavior of IPMC under environmental changes in temperature. This model operated accurately only under small applied voltages close to the operating point. In [32], a generalized Volterra-based model has been presented to model IPMC actuation. The proposed model in [32] is non-autoregressive, which means that no output lags are needed to predict the result. The model achieved tolerable error rate as a non-autoregressive structure. But the main challenge of non-autoregressive modeling of IPMC is to reach the accuracy of a high-fidelity autoregressive model using a non-autoregressive structure. The current paper aims to use an appropriate low-fidelity model to convert the input voltage to some informative data for the identification of IPMC actuation behavior. Then, using converted data and by incorporation of a machine learning (ML) based model, IPMC's behavior can be identified with high accuracy. This hybrid model shows high fidelity to the physical model of IPMC and is more accurate compared to other autoregressive models considering the fact that this hybrid model uses a non-autoregressive structure.

The proposed hybrid model in this study is developed for IPMC soft actuator. This approach can be extended into a wide variety of other applications, no matter whether they are related to IPMC actuator or not. Engineers mostly deal with valid but low-fidelity models. Using the proposed hybrid approach, engineers can use the data acquired from the low-fidelity models to create high-fidelity, interpretable machine learning based models.

The rest of this article is organized as follows: Our proposed hybrid model is introduced in Section II. In section III, the results are discussed, and finally, section IV describes the main motivations of this study as a conclusion.

## II. PROPOSED HYBRID MODEL

In this section, we introduce and substantiate a generalized multimethod approach to model the behavior of non-linear, dynamic, and open-loop systems. The main purpose of using the approach mentioned earlier was to predict the dynamic behavior of the tip displacement of an IPMC actuator given the input voltage. To achieve this goal, we designed a low-fidelity electrical model based on the characteristics of an IPMC actuator. Using this analytical model, a mediatory time series similar to the actual tip displacement was generated, which is the input to the next model. The mediatory signal helps us to precisely predict the actual tip displacement.

Before the advent of numerical methods and machine learning for predicting the behavior of physical systems, models were used to analytically conceptualize the system's behavior; This approach was based on the fact that capturing the physics of the problem, simplified abstract modeling with the capability of providing valuable information could be done to estimate the system's behavior. These models operated well for non-complex systems, but they did not go as expected for modeling non-linear and dynamic systems. Consequently, IPMCs cannot be accurately modeled using only analytical approaches as they exhibit hysteresis and non-linear behavior. Considering this, a hybrid model composed of an RC distributed circuit and a multi-layer perceptron was employed to predict IPMC behaviors. The proposed model can be generalized to other applications and modeling problems.

As illustrated in Fig.2, the proposed hybrid model is comprised of two main models, an analytical model and a machine learning model connected in series. The final goal of this model is to find a non-linear and dynamic mapping between the applied voltage ($V_i(t)$) and the output tip displacement of IPMC ($w(t)$). A conventional method for solving such problems is to directly predict $w(t)$ given $V_i(t)$ using data-driven methods like machine learning models. This approach is not accurate enough when it comes to using non-autoregressive structures. To address this problem, we found that measured voltage on the tip of IPMC ($V_o(t)$) is more correlated to $w(t)$ than $V_i(t)$. Hence, it appears to be a better candidate as the input of the machine learning model of IPMC actuator. In practical applications and especially in large deformation usages, we cannot measure $V_o(t)$ without displacement depreciation of IPMC. This research focuses on the calculation of $V_o(t)$ instead of directly measuring it. To calculate $V_o(t)$, an analytical model based on an RC distributed network was designed which produces output $V_o(t)$. As an example, shown in Fig.3, calculated $V_o(t)$ in comparison to $V_i(t)$ is more similar to $w(t)$. As mentioned earlier, $V_o(t)$ can be more informative as the input for machine learning model of IPMC. Therefore, $V_o(t)$ is fed into the machine learning model which predicts $w(t)$ or the tip displacement accurately without the need for using autoregressive structures. In conclusion, using the proposed hybrid model, an accuracy comparable to a high-fidelity autoregressive model can be achieved using a non-autoregressive structure. In this section, the analytical and the machine learning model are explained in detail.

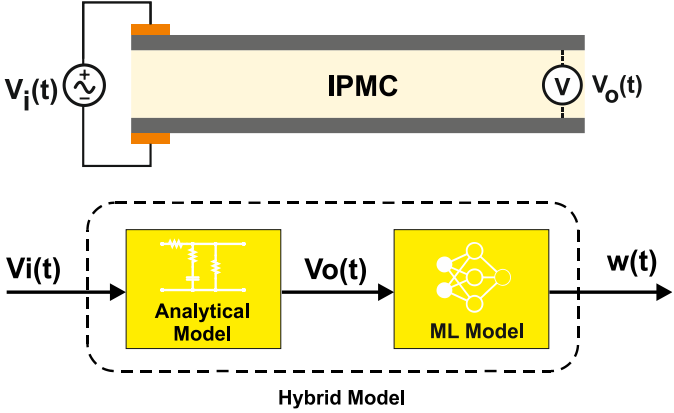

Fig.2. (Up) Representation of Vi(t) and Vo(t) on the body of IPMC. (Down) Block diagram of the proposed hybrid model.

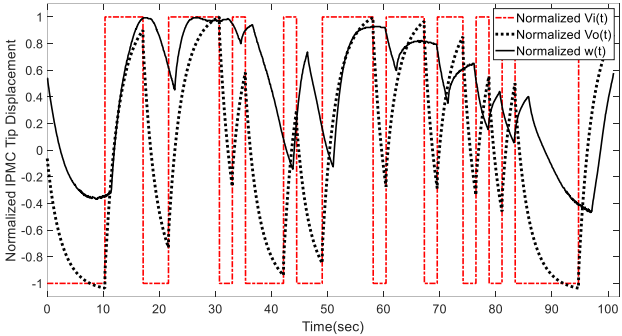

Fig. 3. An example to show that calculated Vo(t) in comparison to Vi(t) is more similar to w(t).

### A. RC Distributed Primary Model

The RC distributed model alone is not accurate enough to model the IPMC tip displacement and is considered a low-fidelity model for describing such a problem. Low-fidelity models cannot precisely reproduce the state and behavior of a system but generally, behave like the actual system. Here, our low fidelity model produces a mediatory signal which is the input to the machine learning model.

As depicted in Fig.4, the proposed equivalent circuit model is comprised of parallel RC stages. Each compartment is comprised of three resistors and one capacitor. $RE_k$ is the k'th compartment modeling electrode resistance. $RI_k$ is the resistance modeling the polymer-metal interface. Also, a capacitance is be formed due to the electrochemical adsorption process at the polymer-metal interface modeled as $C_k$. Finally, $RM$ represents the constant membrane resistance of all compartments. The membrane resistance is deemed constant as the electrical current passing through the membrane is negligible, but $RE_k$, $RI_k$, and $C_k$ (Eqs 4-6) increase linearly with regard to compartment number. This increase represents the attenuation occurring in the electrical current passing through the actuator.

$$RE_k = \frac{(RE_{tip} - RE_{clamp})k}{N} + RE_{clamp} \qquad (4)$$

$$RI_k = \frac{(RI_{tip} - RI_{clamp})k}{N} + RI_{clamp} \qquad (5)$$

$$C_k = \frac{(C_{tip} - C_{clamp})k}{N} + C_{clamp} \qquad (6)$$

N in (Eqs 4-6) is the total number of compartments and indexes of *tip* and *clamp* refer to resistance and capacitance of the tip and clamping area of IPMC. For example, $RE_{tip}$ and $RE_{clamp}$ refer to electrode resistance of IPMC in tip and clamping regions. (Eqs 7-9) describe the change in resistance or the capacitance of the IPMC actuator tip compared with respect to the initial value measured in the clamping region. The ratio of the measured parameter at the clamping region to the measured parameter at the tip of the actuator is called the attenuation coefficient ranging from 0 to 1 ($\alpha_E, \alpha_I, \alpha_C$).

$$RE_{tip} = (\alpha_E^{-1}) RE_{clamp} \qquad (7)$$

$$RI_{tip} = (\alpha_I^{-1}) RI_{clamp} \qquad (8)$$

$$C_{tip} = (\alpha_C^{-1}) C_{clamp} \qquad (9)$$

Where $\alpha_E$ is the current attenuation coefficient on the surface electrodes. $\alpha_I$ is the current attenuation coefficient in the interface of the membrane and the electrode. Finally, $\alpha_C$ is the capacitance attenuation coefficient.

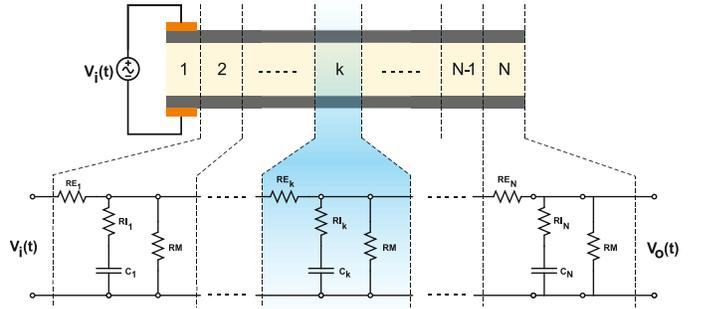

Fig. 4. Proposed RC distributed network for IPMC.

Each compartment of the proposed network is a single linear and time-invariant (LTI) RC circuit. Hence, we are allowed to calculate the Laplace transforms of each stage to find its transfer function. For instance, $V_k(s)$ and $V_{k+1}(s)$ represent the input and output voltages of the k'th compartment (Fig.5). Also $H_k(s)$ is the transfer function of the k'th compartment and is defined as follows:

Eq 10 gives the transfer function of a single stage RC distributed network.

$$H_k(s) = \frac{V_{k+1}(s)}{V_k(s)} = G_k \frac{s+Z_k}{s+P_k} \quad (10)$$

Where $s$ is the complex variable of Laplace transform and $G_k$, $Z_k$ and $P_k$ are the gain, zero, and pole of the k'th stage, obtained from the following equations:

$$G_k = \left(1 + \frac{RE_k}{RM} + \frac{RE_k}{RI_k}\right)^{-1} \quad (11)$$

$$Z_k = (RI_k C_k)^{-1} \quad (12)$$

$$P_k = \frac{1}{C_k}\left(RI_k + \frac{RE_k RM}{RE_k + RM}\right)^{-1} \quad (13)$$

Based on the (Eqs 10-13), the transfer function of each compartment can be obtained. The total transfer function of the network is calculated by multiplication of all the transfer functions of the N stages as depicted in Fig.6. $H(s)$ is the system's transfer function and represents the relationship between $V_i(s)$ and $V_o(s)$ which is defined as follows:

$$H(s) = \frac{V_o(s)}{V_i(s)} = \prod_{k=0}^{N}\left(G_k \frac{s+Z_k}{s+P_k}\right) \quad (14)$$

By applying the inverse Laplace transform on the transfer function given in Eq 14, the output voltage measured in the tip of the IPMC actuator in the time domain ($V_o(t)$) can be calculated in Eq 15.

$$V_o(t) = V_i(t) * h(t) = \int_0^t V_i(\tau)h(t-\tau)d\tau \quad (15)$$

Where '*' refers to the convolution operator and $h(t)$ is the time domain version of H(s).

$$h(t) = h_1(t) * \cdots * h_k(t) * \cdots * h_N(t) \quad (16)$$

And,

$$h_k(t) = -P_k Z_k \left(\int_0^t e^{-P_k \tau} e^{-P_k(t-\tau)} d\tau + te^{-P_k t}\right) \quad (17)$$

To obtain the actual value of $V_o(t)$ some of the geometrical (Fig.7) and physical parameters of the metal electrode, polymer membrane, and the polymer-metal interface are required. These parameters are shown in Table. 1. $L$, $W$, $h$, $A$, $\rho$, and $\sigma$ refer to the length, width, thickness, electric resistivity, and electrical conductivity respectively that all are known. $\xi_\rho$ and $\xi_h$ are two unknown parameters that refer to Interface Metal Resistance Ratio (IMRR) and Interface Metal Thickness Ratio (IMTR), respectively, obtained during the parameter estimation process.

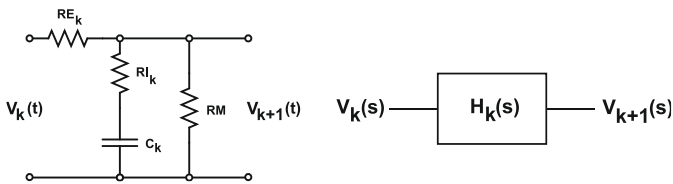

**Figure. 5. (Left): The k'th single RC compartment. (Right): Block diagram of the k'th single RC compartment which $H_K(s)$ is the transfer function this compartment.**

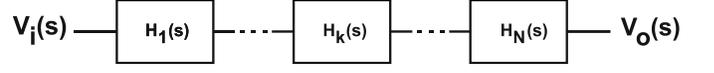

**Fig.6. Block diagram of a cascaded N stage model of an IPMC actuator**

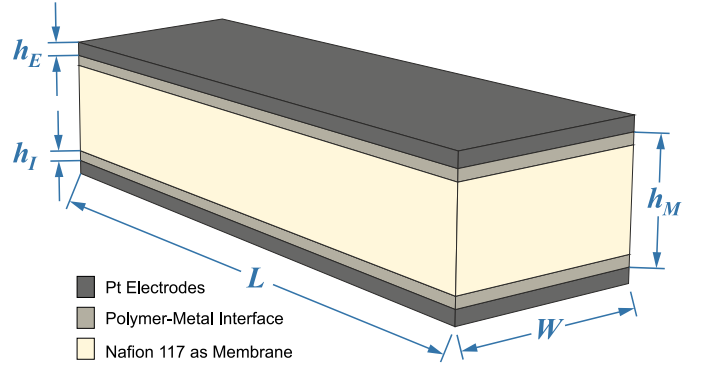

**Figure. 7. An IPMC actuator.**

| Metal Electrode (Pt) | |
|---|---|
| $h_E$ | $10^{-6}$ (m) |
| $\rho_E$ | $1.06 \times 10^{-7}$ (Ω.m) |
| $A_E = W.h_E$ (m²) | |
| $RE_{clamp} = \rho_E \frac{L}{A_E}$ (Ω) | |
| **Polymer Membrane (Nafion® 117)** | |
| $h_M$ | $183 \times 10^{-6}$ (m) |
| $\sigma_M$ | 10.26 (℧/m) |
| $\rho_M = 1/\sigma_M$ (Ω.m) | |
| $A_M = W.h_M$ (m²) | |
| $R_M = \rho_M \frac{L}{A_M}$ (Ω) | |
| **Polymer-Metal Interface** | |
| $h_I = \xi_h . h_E$ (m) | |
| $\rho_I = \xi_\rho . \rho_E$ (Ω.m) | |
| $A_I = W.h_I$ (m²) | |
| $RI_{clamp} = \rho_I \frac{L}{A_I}$ (Ω) | |
| **Dimensions** | |
| L | 22 (mm) |
| W | 5.5 (mm) |

### B. Machine Learning Based Complementary Model

As a complementary model, a deep multi-layer perceptron (D-MLP) neural network was designed. The network architecture is composed of 11 fully connected hidden layers. Each layer of the network is comprised of 10 neurons. According to the universal approximation theorem, MLPs with at least one hidden layer can approximate any continuous function. Hence, they are suitable for being used as a complementary model for processing the mediatory signal represented as time series. The network architecture is represented in Fig.8. As mentioned earlier, a mediatory signal ($V_o(t)$) is generated by the analytical model. The mediatory signal, which is more similar to the tip displacement of the IPMC actuator compared to the input stimulus, needs to be restructured so that the problem can be solved with the help of supervised learning approaches. Framing the time series by fixed windows is a conventional approach in dealing with time series classification and forecasting problems. In this method, a fixed-length sliding window of size $\tau$ slides over the time series to reshape the data into input series of size $\tau$. In other words, the model is mapping the previous $\tau$ data points from time step t- $\tau$ to t to predict tip displacement of IPMC at $t$ (w(t)). In this work, 30 frames of the data were recorded per second, and the experiments showed that keeping the information from the previous 2 seconds of the IPMC behavior resulted in acceptable performance, so each prediction takes 60 evenly spaced data points as the input. Every two consecutive datapoints are 2/60 seconds (~33.33 ms) apart. The dataset was split into train, test, and validation data by the ratio of 30%, 50% and 20%. Also, Levenberg-Marquardt backpropagation training algorithm was utilized to determine the synaptic weights of the network.

## III. RESULTS AND DISCUSSION

In order to evaluate the proposed model, a hardware setup is required to record IPMC's tip displacement. In this section, the hardware setup used will be introduced first, then the experimental results are presented.

### A. Hardware Apparatus

The hardware setup, shown in Fig.9 , cis comprised of two primary parts:
- Actuation setup:

    This segment aims to create and apply an input stimulus voltage to the IPMC actuator. The desired voltage is created digitally on MATLAB. The voltage is sent to an Arduino UNO board via Arduino support package on MATLAB. Also, we used a 12-bit Digital to Analog Converter (DAC) to generate an analog output voltage. The output voltage generated by Arduino is then shifted and amplified using an analog circuit containing filters and differential amplifiers has been used. This voltage can now be applied to the IPMC.

- Capturing setup:

    IPMC actuator should be held by a tong firmly during each trial of the test. Also, the gripping position of the IPMC actuator should remain fixed during the entire experiment. Hence, two holders were assembled for holding the IPMC actuator and the camera. Here, a video captures IPMC's position, which will be further processed to record the tip position of the actuator. The sampling rate of the camera used was 30 fps which remained constant during the experiment.

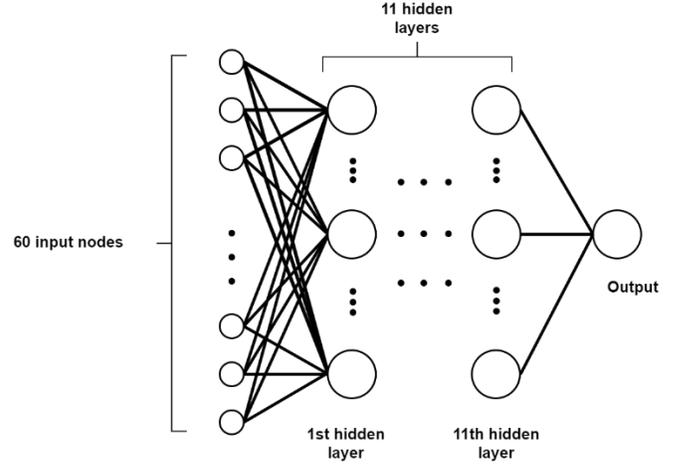
Fig. 8 The proposed D-MLP Neural network.

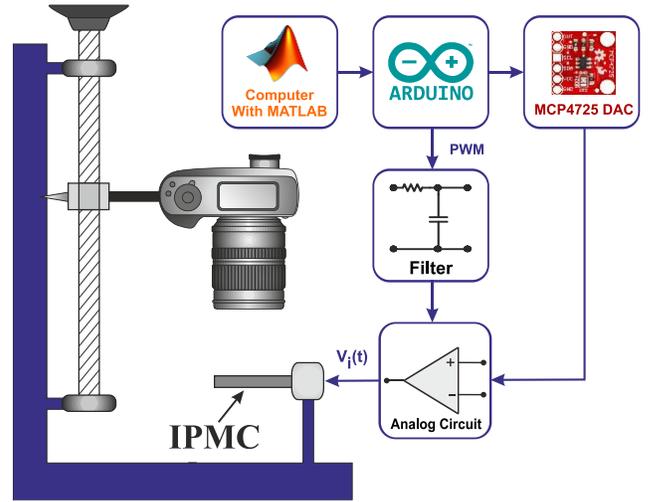
Figure. 9. Hardware apparatus.

### B. Experimental Results

As illustrated in Fig.2, The model consists of two main parts. The physical model is a distributed RC circuit representing the IPMC's behavior using 45 RC stages (N=45). Machine learning model is a time-delayed dynamic deep MLP with a two-second memory. To examine the necessity of the existence of an analytical model, two different models are tested out. The first model, only uses a deep neural network to predict the tip displacement of the IPMC actuator. The network used in this test is exactly the same as the network used in the second experiment. The second model is the proposed hybrid model including both analytical and intelligent models with the properties mentioned earlier. Both models have been trained using four different input voltages: PRBS, Sine, Chirp, and Pulse. After training each individual model, two performance measures have been calculated to evaluate the two structures. The first measure is Normalized Mean Square Error (NMSE), a performance criterion commonly used to evaluate how well the model can predict the output. Eq.18 shows how this criterion is calculated.

$$NMSE\big(w(t),\hat{w}(t)\big) = MSE\big(w_n(t),\hat{w}_n(t)\big) \qquad (18)$$

$\hat{w}(t)$ is the estimated value for IPMC's tip displacement. $w_n(t)$ and $\hat{w}_n(t)$ refer to normalized values of $w(t)$ and $\hat{w}(t)$ calculated by dividing each value of w(t) and $\hat{w}(t)$ by the maximum values observed. Also, MSE or Mean Squared Error is calculated as follows:

$$MSE(w_n(t), \hat{w}_n(t)) = \frac{1}{M} \sum_{All} (w_n(t) - \hat{w}_n(t))^2 \quad (19)$$

Where $M$ is the length of $w(t)$.

The second criterion used is called Fitting, which shows how well the model fits the shape of output. It can be calculated as:

$$Fitting = 100 \left( 1 - \frac{\|w_n(t) - \hat{w}_n(t)\|}{\|w_n(t) - mean(w_n(t))\|} \right) \quad (20)$$

The results are displayed in Figs 10 to 13. These figures show the IPMC's tip displacement over the time (as Target) for four different signals introduced earlier and the output results of two mentioned models (Normal and Hybrid models). Obviously the desire of the model is fully tracking of Target. Furthermore, both models have been evaluated with two above mentioned criteria. Its result is presented in Table 2.

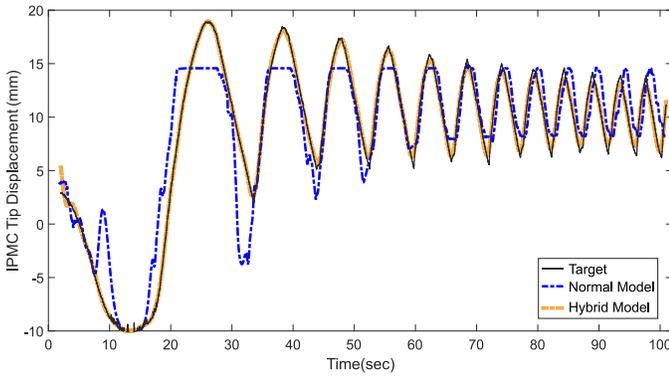

Fig.10 . The actual and estimated (using Normal and Hybrid models) tip displacements of IPMC in response to a chirp input voltage.

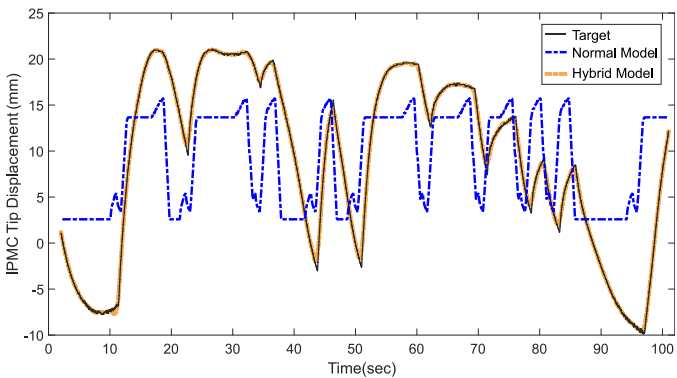

Fig.11 . The actual and estimated (using Normal and Hybrid models) tip displacements of IPMC in response to a PRBS input voltage.

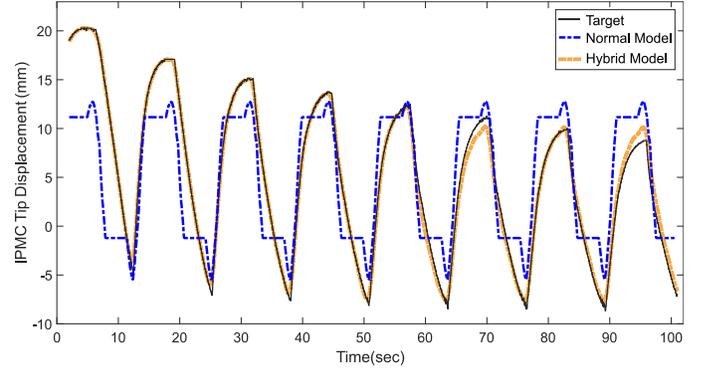

Fig.12 . The actual and estimated (using Normal and Hybrid models) tip displacements of IPMC in response to a Pules train input voltage.

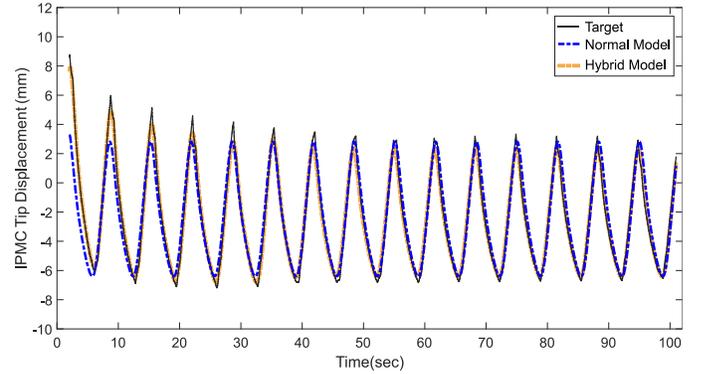

Fig.13 . The actual and estimated (using Normal and Hybrid models) tip displacements of IPMC in response to a sine input voltage.

Table. 2. Evaluating the Hybrid model and the Normal model with NMSE and Fitting. H represent the Hybrid model and N shows the Normal model

|  | NMSE | | Fitting | |
| --- | --- | --- | --- | --- |
|  | H | N | H | N |
| **PRBS** | 1.1225e-04 | 0.1595 | %97.71 | %16.62 |
| **Sine** | 0.0027 | 0.5983 | %92.27 | %71.79 |
| **Chirp** | 2.64e-04 | 0.0375 | %95.97 | %58.37 |
| **Pulse** | 7.55e-04 | 0.0694 | %92.90 | %42.79 |
| **Average** | 9.5781e-04 | 0.2162 | %94.71 | %47.39 |

As mentioned earlier, few research has been done on non-autoregressive models. Comparing the performance measures resulted in this study to the results achieved in previous works, it can be easily concluded that the proposed hybrid model performs much better. For instance, comparing the results achieved from non-autoregressive Volterra-based model introduced in [32] to the hybrid model proposed in this study shows drastic declines in NMSE values. The NMSE value reported in [32] was 0.0692 while using the hybrid model introduced in this study, the NMSE reached to a value of 9.5781e-04. The results of the proposed non-autoregressive model are even comparable with autoregressive structures in which information from the previous time stamps of the output are required for making predictions. This indicates that the proposed open-loop model can accurately predict the output with an accuracy comparable with the closed loop ones. For example, the average NMSE error reached using ANFIS-NARX method in [28] is 5.86e-5 which is close to the results reported Table. 2. While our proposed hybrid model is non-autoregressive but the ANFIS-NARX paradigm of [28] uses an autoregressive structure. On the other hand, the average fitting of %94.71 for the hybrid model shows that this model can describe IPMC's actuation behavior highly precise. While the same neural network structure only reaches an average fitting of %47.39. According to figures 10 to 13, it can also be observed that the machine learning model on its own showed less accuracy than the hybrid model in which the analytical model is combined with a MLP structure. It is clear that the neural network alone does not reach tolerable error rates. This demonstrates the usefulness of using hybrid models in which a mediatory signal is generated by the analytical sub-model which is then fed into the machine learning model.

## IV. CONCLUSION

In this paper, we introduced a hybrid analytical intelligent model for non-linear, dynamic, and open-loop identification of IPMC soft artificial muscles using a multimethod approach. This multimethod approach was tested and validated by estimating the voltage-displacement relation of an IPMC artificial muscle actuator. According to the results shown in section III, the proposed model performs much better in comparison to the previous works. The model uses an open-loop structure, eliminating the need for continuous feedback to reduce the error signals. Using physical, geometrical, and mechanical properties of an already fabricated IPMC actuator, we designed an analytical and interpretable electrical circuit model of the sensor using which a mediatory signal similar to the tip displacement was generated. The mediatory signal was then fed into a deep MLP network to generate the actual tip displacement of the fabricated IPMC.

The electrical circuit modeled here is comprised of RC distributed compartments. This model is an example of a low fidelity model of a non-linear system. Low-fidelity models are not appropriate for precisely modeling the behavior of non-linear systems, but they can produce roughly correct results that can be used as readable and interpretable data that can be further analyzed. This is the primary motivation to use a multimethod approach. This approach is not restricted to predicting IPMC actuators behaviors and can be used in other applications where analytical models can be built.